\documentclass[10pt,twocolumn,letterpaper]{article}

\usepackage{iccv}
\usepackage{times}
\usepackage{epsfig}
\usepackage{graphicx}
\usepackage{amsmath}
\usepackage{amssymb}

\DeclareMathOperator*{\argmin}{arg\,min}
\usepackage{algpseudocode}
\usepackage{algorithm}
\usepackage{array}
\usepackage{multirow}
\usepackage{dsfont}
\usepackage[font={small}]{caption}
\usepackage{wrapfig,lipsum,booktabs}

\usepackage{color}
\usepackage{xcolor}
\definecolor{citecolor}{HTML}{0071bc}
\usepackage[pagebackref=true,breaklinks=true,letterpaper=true,colorlinks,citecolor=citecolor,bookmarks=false]{hyperref}

\iccvfinalcopy 

\def\iccvPaperID{****} 

\newlength\savewidth\newcommand\shline{\noalign{\global\savewidth\arrayrulewidth
  \global\arrayrulewidth 1pt}\hline\noalign{\global\arrayrulewidth\savewidth}}
\newcommand{\tablestyle}[2]{\setlength{\tabcolsep}{#1}\renewcommand{\arraystretch}{#2}\centering\footnotesize}

\begin{document}

\title{A-SDF: Learning Disentangled Signed Distance Functions \\ for Articulated Shape Representation}

\author{Jiteng Mu\textsuperscript{1}, \quad
Weichao Qiu\textsuperscript{2}, \quad
Adam Kortylewski\textsuperscript{2}, \quad
Alan Yuille\textsuperscript{2}, \\
Nuno Vasconcelos\textsuperscript{1}, \quad
Xiaolong Wang\textsuperscript{1}\\
\textsuperscript{1}UC San Diego, \textsuperscript{2}Johns Hopkins University
}

\twocolumn[{%
\vspace{-1em}
\maketitle
\vspace{-1em}

\begin{center}
    \centering
    \vspace{-0.2in}
    \includegraphics[width=\linewidth]{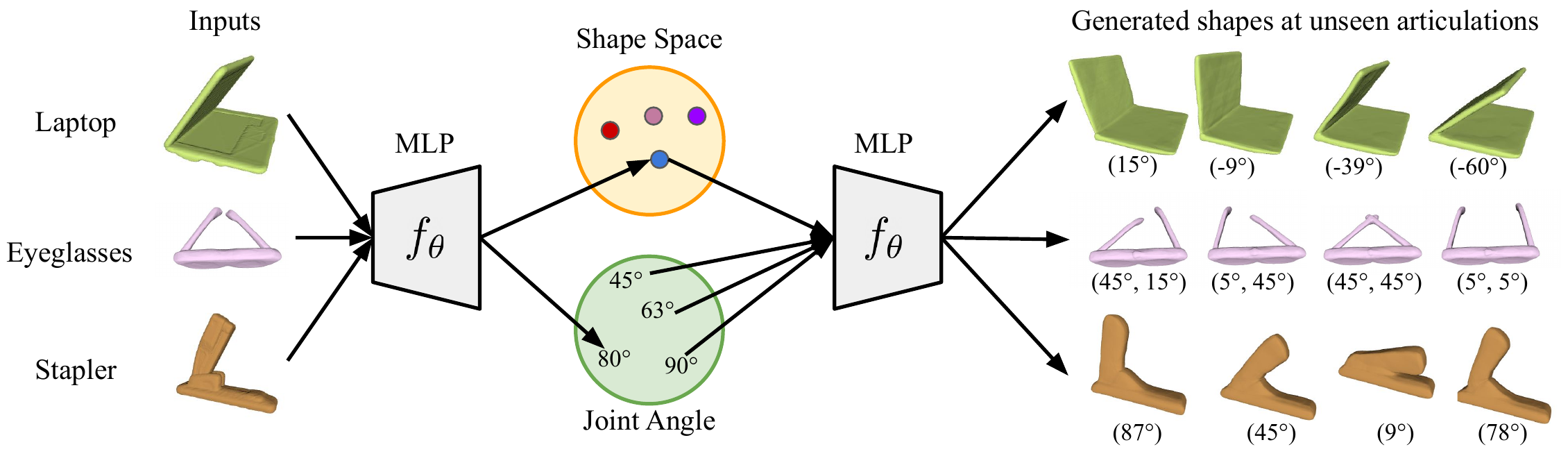}
    \vspace{-0.2in}
    \captionof{figure}{
    We represent articulated objects with separate codes for encoding shape and articulation. During inference, given an unseen instance, our model first infers the shape and articulation codes via back-propagation. With the inferred shape code, we can generate shapes at unseen angles by only changing the articulation code.}
    \label{fig:teasing}
\end{center}
}]



\begin{abstract}
\vspace{-0.1in}
Recent work has made significant progress on using implicit functions, as a continuous representation for 3D rigid object shape reconstruction. However, much less effort has been devoted to modeling general articulated objects. Compared to rigid objects, articulated objects have higher degrees of freedom, which makes it hard to generalize to unseen shapes. To deal with the large shape variance, we introduce Articulated Signed Distance Functions (A-SDF) to represent articulated shapes with a disentangled latent space, where we have separate codes for encoding shape and articulation. We assume no prior knowledge on part geometry, articulation status, joint type, joint axis, and joint location. With this disentangled continuous representation, we demonstrate that we can control the articulation input and animate unseen instances with unseen joint angles. Furthermore, we propose a Test-Time Adaptation inference algorithm to adjust our model during inference. We demonstrate our model generalize well to out-of-distribution and unseen data, e.g., partial point clouds and real-world depth images.
Project page: \href{https://jitengmu.github.io/A-SDF/}{https://jitengmu.github.io/A-SDF/}.
\end{abstract}
\vspace{-0.15in}
\section{Introduction}

Modeling articulated objects has wide applications in multiple fields including virtual and augmented reality, object functional understanding, and robotic manipulation. To understand articulated objects, recent works propose to train deep networks for estimating per-part poses and the joint angle parameters of an object instance in a known category~\cite{DBLP:conf/cvpr/Li0YGAS20,weng2021captra}. However, if we want to interact with the articulated object (e.g., open a laptop), estimating its static state is not sufficient. For example, an autonomous agent needs to predict what the articulated object shape will be like after interactions for  planning its action. 


In this paper, we introduce Articulated Signed Distance Functions (A-SDF), a differentiable category-level articulated object representation, which can reconstruct and predict the object 3D shape under different articulations. A differentiable model is useful in applications which require back-propagation through the model to adjust inputs, such as rendering in graphics and model-based control in robotics. 

We build our articulated object model based on the deep implicit Signed Distance Functions ~\cite{DBLP:conf/cvpr/ParkFSNL19}. While implicit functions have recently been widely applied in modeling static object shape with fine details~\cite{DBLP:conf/iccv/SaitoHNMLK19,DBLP:conf/cvpr/SaitoSSJ20,DBLP:conf/nips/SitzmannZW19}, much less effort has been devoted to modeling general articulated objects. This is an extremely challenging task, as we assume no prior knowledge on part geometry, articulation status, joint type, joint axis and joint location about the object class. Considering the possible combinations of the above factors, even injecting one more degree of freedom into an object raises the modeling space a magnitude up. We observe that models with a single shape code input, such as DeepSDF~\cite{DBLP:conf/cvpr/ParkFSNL19}, cannot encode the articulation variation reliably. It is even harder for the models to generalize to unseen instances with unseen joint angles.

To improve the generalization ability, we propose to model the joint angles explicitly for articulated objects. Instead of using a single code to encode all the variance, we propose to use one shape code to model the shape of object parts and a separate articulation code for the joint angles. To achieve this, we design two separate networks in our model: (i) a shape encoder to produce a shape embedding given a shape code input; (ii) an articulation network which takes input both the shape embedding and an articulation code to deform the object shape. During training, we use the ground-truth joint angles as inputs and learn the shape code jointly with both model parameters. To enable the disentanglement, we enforce the same instance with different joint angles to share the same shape code.

During inference, given an unseen instance with unknown articulation, we first infer the shape code and articulation code via back-propagation. Given the inferred shape code, we can simply adjust the articulation code to generate the instance at different articulations. We visualize the generation process and results for a few objects in Figure~\ref{fig:teasing}. Note the part geometry remains the same as we fix the inferred shape code during generation.

To generalize our model to out-of-distribution and unseen data, e.g., partial point clouds and real-world depth images, we further propose a Test-Time Adaptation (TTA) approach to adjust our model during inference. Note that our unique model architecture with separate shape encoder and articulation network provides the opportunity to do so: As the separation of shape encoder and articulation network ensures the disentanglement is maintained when the shape encoder is adapted. We adapt the shape encoder network to the current test instance by updating its parameters, while fixing the parameters of the articulation network. This procedure allows A-SDF to reconstruct and generate better shapes aligning with the inputs. 



To our knowledge, our work is the first paper tackling the problem of generic articulated object synthesis in the implicit representation context. We summarize the contributions of our paper as follows. First, we propose Articulated Signed Distance Functions (A-SDF) and a Test-Time Adaptation inference algorithm to model daily articulated objects. Second, the disentangled continuous representation allows us to control the articulation code and generate corresponding shapes as output on unseen instances with unseen joint angles. Third, the proposed representation shows significant improvement on interpolation, extrapolation, and shape synthesis. More interestingly, our model can generalize to real-world depth images from the RBO dataset~\cite{DBLP:journals/ijrr/Martin-MartinEB19} and we quantitatively demonstrate superior performance over the baselines. 
\section{Related Work}

\textbf{Neural Shape Representation.} A large body of work~\cite{DBLP:conf/eccv/WangZLFLJ18,DBLP:conf/iccv/Gkioxari0M19,DBLP:conf/cvpr/ChenTZ20,DBLP:conf/cvpr/DengGYBHT20,DBLP:conf/cvpr/LiaoDG18,DBLP:conf/cvpr/RieglerUG17,DBLP:journals/corr/abs-1912-06126,DBLP:conf/eccv/VarolCRYYLS18,DBLP:conf/cvpr/TulsianiSGEM17,DBLP:conf/iccv/ZouYYCH17,DBLP:conf/nips/DeprelleGFKRA19,DBLP:conf/iclr/AchlioptasDMG18,DBLP:conf/cvpr/FanSG17} has focused on investigating efficient and accurate 3D object representations. Recent advances suggest that representing 3D objects as continuous and differentiable implicit functions~\cite{DBLP:conf/cvpr/GenovaCSSF20,DBLP:conf/cvpr/ParkFSNL19,DBLP:conf/cvpr/MeschederONNG19,DBLP:conf/cvpr/ChenZ19,DBLP:journals/corr/abs-2003-12673,DBLP:conf/cvpr/JiangSMHNF20,DBLP:conf/nips/XuWCMN19,DBLP:conf/nips/SitzmannZW19,DBLP:conf/cvpr/NiemeyerMOG20,DBLP:conf/iccv/SaitoHNMLK19,DBLP:conf/eccv/MildenhallSTBRN20,DBLP:journals/corr/abs-2006-09661,DBLP:journals/corr/abs-2010-04595} can model various topologies in a memory-efficient way. The basic idea is to exploit neural networks to parameterize a shape as a decision boundary in 3D. 
Most of these work is limited to modeling static objects and scenes~\cite{DBLP:conf/cvpr/GenovaCSSF20,DBLP:conf/cvpr/JiangSMHNF20,DBLP:conf/nips/XuWCMN19,DBLP:conf/nips/SitzmannZW19,DBLP:conf/cvpr/NiemeyerMOG20,DBLP:conf/iccv/SaitoHNMLK19,DBLP:conf/eccv/MildenhallSTBRN20,DBLP:journals/corr/abs-2006-09661,DBLP:journals/corr/abs-2010-04595}. Different from previous works, our method models articulated objects in a category-level by learning a disentangled implicit representation and we test our model on real depth images. Comparisons to implicit neural networks on deformable shapes will be discussed in depth in the following.

\textbf{Articulated Humans.} One line of work leverages parametric mesh models~\cite{DBLP:journals/tog/LoperM0PB15,DBLP:journals/tog/LiBBL017,DBLP:conf/cvpr/ZuffiKJB17,bogo2016keep} to estimate shape and articulation for faces~\cite{DBLP:conf/cvpr/TranHMM17,DBLP:conf/eccv/RanjanBSB18,DBLP:conf/cvpr/SanyalBFB19}, hands\cite{DBLP:conf/cvpr/GeRLXWCY19}, humans bodies~\cite{peng2021neural,DBLP:conf/iccv/BhatnagarTTP19,DBLP:conf/iccv/ZhengYWDL19,DBLP:conf/cvpr/KocabasAB20,DBLP:conf/cvpr/KanazawaBJM18,DBLP:conf/3dim/OmranLPGS18,DBLP:conf/eccv/ZhangPJRMK20}, and animals~\cite{DBLP:conf/cvpr/ZuffiKB18,kanazawa2016learning,kulkarni2020articulation,zuffi2019three} by directly inferring shape and articulation parameters.
However, such parametric models requires substantial efforts from experts to construct and thus is hard to generalize to large-scale object categories. To address the challenge, another line of work~\cite{pablo2021npms,DBLP:conf/iccv/NiemeyerMOG19,DBLP:conf/eccv/TretschkTGZST20,DBLP:conf/eccv/DengLJPH0T20,xu2021snarf,noguchi2021narf} employs neural networks to learn shapes from data. 
For example, Niemeyer et al.~\cite{DBLP:conf/iccv/NiemeyerMOG19} learned an implicit vector field assigning every point with a motion vector and deformed shapes in a spatial-temporal space. However, the design does not allow for controlling each part separately. Recently, Hoang et al.~\cite{DBLP:conf/eccv/TretschkTGZST20} defined shapes using patches and the overall shape can be changed by manipulating the defined extrinsic parameters of each patch. Nevertheless, the learned patches do not correspond to parts and the method fails at large deformation. Deng et al.~\cite{DBLP:conf/eccv/DengLJPH0T20} modeled each human body part by a separate implicit function. However, the method is limited to instance-level and requires skinning weights for the learning process. In comparison, our method is category-level on general articulated objects and we assume no part label. Therefore, these previous approaches are not directly comparable to ours. Besides, we model articulated object poses with joint angles, which allows us to articulate each joint separately.

\textbf{Articulated General Objects.} Though reconstructing articulated humans has attracted lots of attention in the community, modeling 3D shapes of daily articulated objects is an under-explored field in terms of both data and approaches. Unlike modeling humans where lots of human priors are injected, modeling articulated daily objects poses additional challenges by assuming no prior part geometry and labels, articulation status, joint type, joint axis and joint location about the category. One recent popular paradigm of research~\cite{DBLP:conf/icra/HausmanNOS15,DBLP:conf/icra/KatzB08,DBLP:conf/icra/MartinHB16,DBLP:conf/icra/DesinghLOJ19,DBLP:conf/bmvc/MichelKBYGR15,weng2021captra} focuses on estimating 6D poses of articulated objects. For example, Desingh et al.~\cite{DBLP:conf/icra/DesinghLOJ19} proposed a instance-level factored approach to estimate the poses of articulated objects with articulation constraints. 
However, 6D pose information may not be sufficient for tasks that require detailed shape information, such as robotic manipulation~\cite{articulated08,Xiang_2020_CVPR,mo2021where2act,mittal2021articulated}. In this work, we show that implicit functions are suitable for the daily articulated objects modeling. We demonstrate that once an implicit function is learned, shapes at unseen articulations can be generated by manipulating the articulation code.

\textbf{Disentangled Representation.} Disentangled representations focus on modeling complex variations in a low dimensional space, where individual factors control different types of variation. Previous work~\cite{kulkarni2015deep,higgins2016beta,chen2016infogan,huang2018munit,kim2018disentangling,karras2019style,DBLP:conf/iclr/HigginsMPBGBML17,DBLP:conf/iccv/Nguyen-PhuocLTR19,DBLP:journals/corr/abs-2007-02442,DBLP:conf/iccv/ChanGZE19,DBLP:conf/nips/ZhuZZ00TF18,DBLP:conf/eccv/ZhouBP20,shen2020interpreting,pidhorskyi2020adversarial,liu2020factorize,anokhin2020high,park2020swapping} has shown that disentangled representations are essential to learn meaningful latent space for 2D image synthesis.
For example, Zhou et al.~\cite{DBLP:conf/eccv/ZhouBP20} proposed an auto-encoder architecture to disentangle human pose and shape. 
Different from previous work, we focus on modeling general articulated objects in 3D with interpretable pose codes.

\textbf{Adaptation on Test Instance.} Learning on test instance has been recently applied for adapting a trained model to out-of-distribution in multiple applications, including image  recognition~\cite{DBLP:conf/cvpr/JainL11,Mullapudi_2019,DBLP:conf/icml/SunWLMEH20}, super-resolution and synthesis~\cite{DBLP:conf/cvpr/ShocherCI18,shocher2018ingan,Bau2019PhotoManipuation}, and mesh reconstruction and generation~\cite{DBLP:conf/nips/LiLMKW0K20,jiang2021hand}. For example, Li et al.~\cite{DBLP:conf/nips/LiLMKW0K20} exploits training in test-time with self-supervision for consistent mesh reconstruction in a single video. Inspired by previous works, we develop a Test-Time Adaptation inference algorithm in implicit functions for better shape reconstruction and generation, where our disentanglement-based model architecture is the key to allow for Test-Time Adaptation. 
\section{Method}

\begin{figure*}
    \centering
    \includegraphics[width=0.95\linewidth]{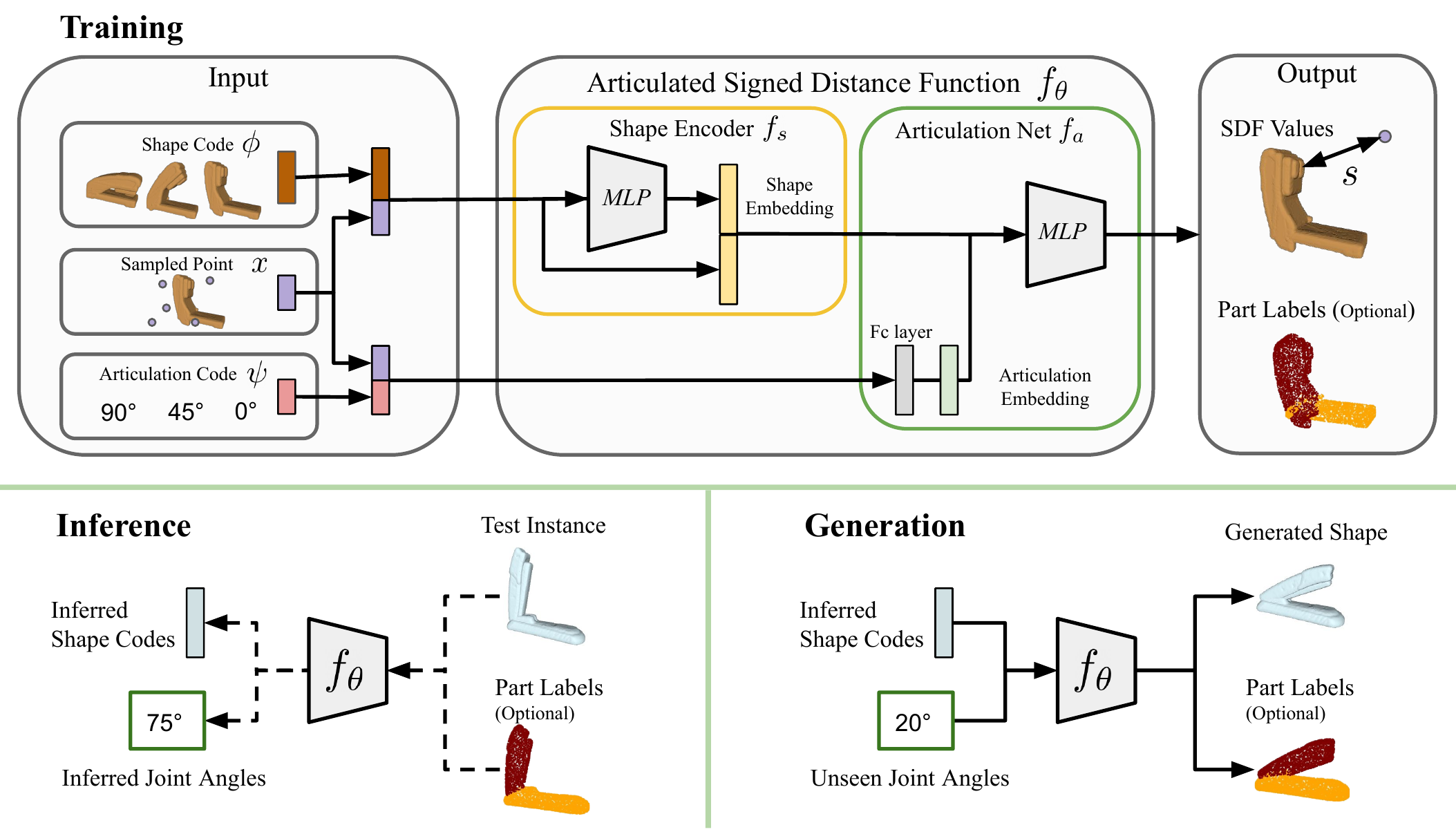}
    \caption{Overview of the proposed method. At training time, the articulation code and randomly sampled shape codes are first concatenated with a sampled point separately. The produced embeddings are then input to an articulated signed distance function $f_{\theta}$ to regress SDF values (signed distance) and predict part labels (optional). Note that the same instance is associated with one shape code regardless of its articulation state, as illustrated with brown staplers. During inference, back-propagation is used to jointly infer the shape code and articulation code for an unseen instance. With the inferred shape code, the model can faithfully generate new shape at unseen articulations. Best viewed in color.}
    \label{fig:overview}
    \vspace{-0.20in}
\end{figure*}

We propose Articulated Signed Distance Functions (A-SDF), a differentiable category-level  articulated object representation to reconstruct and predict the object 3D shape under  different articulations. Our model takes sampled 3D point locations, shape codes, and articulation codes as inputs, and outputs SDF values (signed distance) that measure the distance of a point to the closest surface point. The key insight is that all shape codes of the same instance should be identical, independent of its articulation. We argue that, even though the shapes look quite different for different articulations of the same instance, a good representation should capture this variability in a low dimensional space, since the part geometry remains unchanged. An overview of our method is presented in Figure~\ref{fig:overview}. As different categories vary a lot in terms of articulation, we train separate networks for each category. 

Our model is based on DeepSDF~\cite{DBLP:conf/cvpr/ParkFSNL19}. DeepSDF is an effective and widely acknowledged baseline and its simple network design allows us to focus on the effectiveness of our key idea rather than designing complex architectures. Note that our approach is generic and it can be applied to other advanced architectures as well. For a fair comparison, our model is designed with similar model size as DeepSDF. Furthermore, compared to feed forward designs~\cite{DBLP:conf/cvpr/MeschederONNG19,DBLP:conf/iccv/SaitoHNMLK19}, the optimization-based shape modeling is naturally compatible with Test-Time Adaptation to address the out-of-distribution data as described in Section~\ref{method:Inference}.  

In the following, we describe how to learn a model encouraging the disentanglement of shape and articulation. We also introduce a Test-Time Adaptation inference technique allowing for generating unseen shapes at unseen articulations with high quality.

\subsection{Formulation}\label{method:Formulation}
Consider a training set of $N$ instance models for one object category. Each instance  is articulated into $M$ poses, leading to a training set of $N \times M$ shapes of the category. Let $\mathcal{X}_{n,m}$ denote the shape articulated from instance $n$ with articulation $m$, where $n \in \{1, \ldots, N\}, m \in \{1, \ldots, M\}$.

Each shape $\mathcal{X}_{n,m}$ is assigned with a shape code $\boldsymbol{\phi}_n \in \mathbb{R}^{C}$, where ${C}$ denotes the latent dimension, and an articulation code $\boldsymbol{\psi}_m \in \mathbb{R}^{D}$ with $D$ denoting the number of DoFs. The shape code $\boldsymbol{\phi}_n$ is shared across the same object instance $n$ across different articulations. During training, we maintain and update one shape code for each instance. We use joint angles to represent the articulation code. For example, the articulation code of a 2-DoF object (e.g., eyeglasses) with both joints articulated to $45^{\circ}$ is $\boldsymbol{\psi}_m = (45^{\circ}, 45^{\circ})$. The joint angle is defined as a relative angle to the canonical pose of the object. 

Let $\boldsymbol{x} \in \mathbb{R}^{3}$ be a sampled point from a shape. For notational simplicity, we omit the subscripts and denote $\boldsymbol{\phi}$ and $\boldsymbol{\psi}$ as the corresponding shape and articulation code of the shape. As shown in Figure~\ref{fig:overview}, an Articulated Signed Distance Function $f_{\theta}$ is finally defined with the auto-decoder architecture, which is composed of a shape encoder $f_{s}$ and an articulation network $f_{a}$,
\begin{equation}\label{eq:model}
f_{\theta}(\boldsymbol{x}, \boldsymbol{\phi}, \boldsymbol{\psi}) = f_{a}[f_{s}(\boldsymbol{\boldsymbol{x}, \phi}), \boldsymbol{\boldsymbol{x}, \psi}] = s,
\end{equation}
where $s \in \mathbb{R}$ is a scalar SDF value (the signed distance to the 3D surface). The sign of the SDF value indicates whether the point is inside (negative) or outside (positive) the watertight surface. The 3D shape is implicitly represented by the zero level-set  $f_{\theta}(\cdot) = 0$.

\subsection{Training}\label{method:Training}

During training, given the ground-truth articulation code $\boldsymbol{\psi}$, sampled points and their corresponding SDF values, the model is trained to optimize the shape code $\boldsymbol{\phi}$ and the model parameters $\theta$. 

The training process is illustrated in Figure~\ref{fig:overview}. The shape code is first concatenated with a sampled point $\boldsymbol{x}$ to form vector of dimension $C+3$ and input to the shape encoder. The output of the shape encoder is then concatenated with the shape code to form a shape embedding. Similarly, the articulation code $\boldsymbol{\psi}$ (joint angles) is first concatenated with a sampled point $\boldsymbol{x}$ to form a $D+3$ dimensional vector. 

Then the articulation network takes the concatenated shape embedding and articulation code to predict the SDF value for the input 3D point. It's worth noting that, at the beginning of the articulation network, a fully connected layer is employed to embed the articulation code into a space of dimension $C+3$. This linear layer ensures a linear correlation between the original articulation code $\boldsymbol{\psi}$ and the resulting articulation embedding. Besides the SDF value, part supervision is optionally provided for training, using different labels to index different object parts. The number of object parts is fixed per category. When part supervision is available, a linear classifier is added to the last hidden layer of the articulation network to simultaneously output the part label.

The training loss functions are defined as following. Let $K$ be the number of sampled points per shape. The function $f_{\theta}$ is trained with the per-point $L_1$ loss function to regress SDF values,
\begin{equation}\label{eq:loss-sdf}
\mathcal{L}^{s}(\mathcal{X},\boldsymbol{\phi},\boldsymbol{\psi}) = \frac{1}{K} \sum_{k=1}^{K} \Big|\Big| f_{\theta}(\boldsymbol{x}_{k}, \boldsymbol{\phi}, \boldsymbol{\psi}) - s_{k} \Big|\Big|_{1},
\end{equation}
where $\boldsymbol{x}_{k} \in \mathcal{X}$ is a point of instance $\mathcal{X}$, $s_{k}$ the corresponding ground-truth SDF value, and $k \in \{1, \ldots, K\}$.

When the object part labels are available, we include a complementary auxiliary part classification loss as,
\begin{equation}\label{eq:loss-part}
\mathcal{L}^{p}(\mathcal{X},\boldsymbol{\phi},\boldsymbol{\psi}) = \frac{1}{K} \sum_{k=1}^{K} \Big[ CE\Big( f_{\theta}(\boldsymbol{x}_{k}, \boldsymbol{\phi}, \boldsymbol{\psi}), p_{k} \Big)\Big],
\end{equation}
where $CE$ denotes the cross-entropy loss and $p_{k}$ is the ground-truth label for the part containing $\boldsymbol{x}_{k}$. Intuitively, the part classification task helps the network disambiguate the object parts.

The full loss $\mathcal{L}(x,\boldsymbol{\phi},\boldsymbol{\psi})$ is defined as,
\begin{equation}\label{eq:loss-full}
\begin{split}
  \mathcal{L}(\mathcal{X},\boldsymbol{\phi},\boldsymbol{\psi}) &= \mathcal{L}^{s}(\mathcal{X},\boldsymbol{\phi},\boldsymbol{\psi}) \\
  &+ \lambda_{p}\mathcal{L}^{p}(\mathcal{X},\boldsymbol{\phi},\boldsymbol{\psi}) + \lambda_{\boldsymbol{\phi}} || \boldsymbol{\phi} ||_{2}^{2}.
\end{split}
\end{equation}

Following~\cite{DBLP:conf/cvpr/ParkFSNL19}, we include a zero-mean multivariate-Gaussian prior per shape latent code $\boldsymbol{\phi}$ to facilitate learning a continuous shape manifold. Part coefficient $\lambda_{p}$ and shape code coefficient $\lambda_{\boldsymbol{\phi}}$ are introduced to balance different losses. We optionally set $\lambda_{p}=0$ when the part labels are not available.

At training time, the shape codes are randomly initialized with a Gaussian distribution at the very beginning of training. Each shape code is then optimized through the training steps and it is shared across all the shapes articulated from the same instance. The articulation codes are constants given from the ground-truths. The objective is to optimize the loss function over all $N \times M$ training shapes, defined as follows, 
\begin{equation}\label{eq:train-update}
\begin{split}
\argmin_{\theta, \boldsymbol{\phi}_n} \sum_{n=1}^{N} \sum_{m=1}^{M} \mathcal{L}(\mathcal{X}_{n,m},\boldsymbol{\phi}_{n},\boldsymbol{\psi}_{m}),
\end{split}
\end{equation}
where $\theta$ is the network parameters. 

\subsection{Inference}\label{method:Inference}

\textbf{Basic Inference.} In the inference stage, illustrated in the Inference Section of Figure~\ref{fig:overview}, an instance $\mathcal{X}$ is given and the goal is to recover the corresponding shape code $\boldsymbol{\phi}$ and the articulation code $\boldsymbol{\psi}$. This can be done by back-propagation. The two codes are initialized randomly, the articulation network parameters are fixed, and the codes are inferred jointly by solving the optimization with the following objective,
\begin{equation}\label{eq:infer-update}
\begin{split}
\hat{\boldsymbol{\phi}}, \hat{\boldsymbol{\psi}} = \argmin_{\boldsymbol{\phi}, \boldsymbol{\psi}} \mathcal{L}(\mathcal{X},\boldsymbol{\phi},\boldsymbol{\psi}).
\end{split}
\end{equation}
However, directly applying Equation~\ref{eq:infer-update} is hard, since both the shape and articulation spaces are non-convex. Gradient based optimization can converge to local minima, producing an inaccurate estimation.

To overcome the challenge, we first use Equation~\ref{eq:infer-update} to optimize both shape and articulation codes as our initial estimation. In practice, we observe that the articulation code usually converges to a good estimate but the inferred shape codes tends to lead to noisy outputs. So the estimated articulation code $\hat{\boldsymbol{\psi}}$ is then kept and the shape code is discarded. In the second step, the shape code is re-initialized, the articulation code is fixed to $\hat{\boldsymbol{\psi}}$, and the optimization is only solved for the shape code $\hat{\boldsymbol{\phi}}$.

\textbf{Test-Time Adaptation Inference.} Though fixing the parameters of decoder $f_{\theta}$ makes sense if the test distribution aligns well with the training distribution, it can be problematic given out-of-distribution observations. To generalize better to out-of-distribution data, the Test-Time Adaptation (TTA) for shape encoder $f_{s}$ is further introduced. It is built on the basic inference procedure with the estimated shape code $\hat{\boldsymbol{\phi}}$ and articulation code $\hat{\boldsymbol{\psi}}$. We fix both estimated codes and finetune the shape encoder $f_{s}$ using the following objective, 
\begin{equation}\label{eq:infer-update-TTA}
\begin{split}
\hat{f_{s}} = \argmin_{f_{s}} \mathcal{L}(\mathcal{X},\hat{\boldsymbol{\phi}},\hat{\boldsymbol{\psi}}),
\end{split}
\end{equation}
where $\hat{\boldsymbol{\phi}}$ and $\hat{\boldsymbol{\psi}}$ are obtained as described in the basic inference. Note that our proposed model architecture is the key for TTA. The separation of shape encoder and articulation network ensures the disentanglement is maintained when the shape encoder is finetuned. After adaptation, the generation ability (Secion~\ref{method:Generation}) is still well preserved. Moreover, in practice, one can easily revert to the non-updated model before adapting to individual instance. 

To this end, the proposed full Test-Time Adaptation Inference Algorithm is summarized in Algorithm~\ref{alg}. Since back-propagation is very efficient, the iterative optimization only introduces negligible additional computation. 

\begin{algorithm}
\caption{ Test-Time Adaptation Inference Algorithm}\label{alg} 
\textbf{Input:} Target shape $\mathcal{X}$. \\ \textbf{Output:} shape code $\hat{\boldsymbol{\phi}}$; articulation code $\hat{\boldsymbol{\psi}}$, updated shape encoder $\hat{f_{s}}$.
\begin{algorithmic}[1]
\State Init {$\boldsymbol{\phi} \sim \mathcal{N}(0,\sigma), \boldsymbol{\psi}$} 
\State $\hat{\boldsymbol{\phi}}, \hat{\boldsymbol{\psi}} = \argmin_{\boldsymbol{\phi}, \boldsymbol{\psi}} \mathcal{L}(\mathcal{X},\boldsymbol{\phi},\boldsymbol{\psi})$\\ \Comment{\textit{Articulation Estimation}}
\State {Re-init $\hat{\boldsymbol{\phi}} \sim \mathcal{N}(0,\sigma)$}
\State $\hat{\boldsymbol{\phi}} = \argmin_{\boldsymbol{\phi}} \mathcal{L}(\mathcal{X},\boldsymbol{\phi},\hat{\boldsymbol{\psi}})$ \Comment{\textit{Shape Code Estimation}}
\State $\hat{f_{s}} = \argmin_{f_{s}} \mathcal{L}(\mathcal{X},\hat{\boldsymbol{\phi}},\hat{\boldsymbol{\psi}})$ \Comment{\textit{Test-time Adaptation}}

\end{algorithmic}
\end{algorithm}

\subsection{Articulated Shape Synthesis} \label{method:Generation}
A main advantage of the proposed disentangled continuous representation is that, once a shape code is inferred, it can be applied to synthesize shapes of unseen instances with unseen joint angles, by simply varying the articulation code. This is shown in Figure~\ref{fig:overview}, Generation section. In this stage, the shape code and finetuned shape encoder $f_{s}$ obtained in the inference stage is fixed and new shapes are generated by simply inputting new joint angles to the network. If trained with part labels, the model can also output part labels for unseen shapes, as shown in Figure~\ref{fig:part}.

\section{Experiment}
We first introduce our experimental setup. We then experiment with 3D reconstruction and demonstrate that the proposed method can successfully interpolate, extrapolate the articulation space, and synthesize new shapes from Section~\ref{exp:recon} to Section~\ref{exp:gen}. Finally, we quantitatively show the proposed method generalize well to partial point clouds obtained from both synthetic and real depth images.

\begin{figure}[t]
    \centering
    \includegraphics[width=\linewidth]{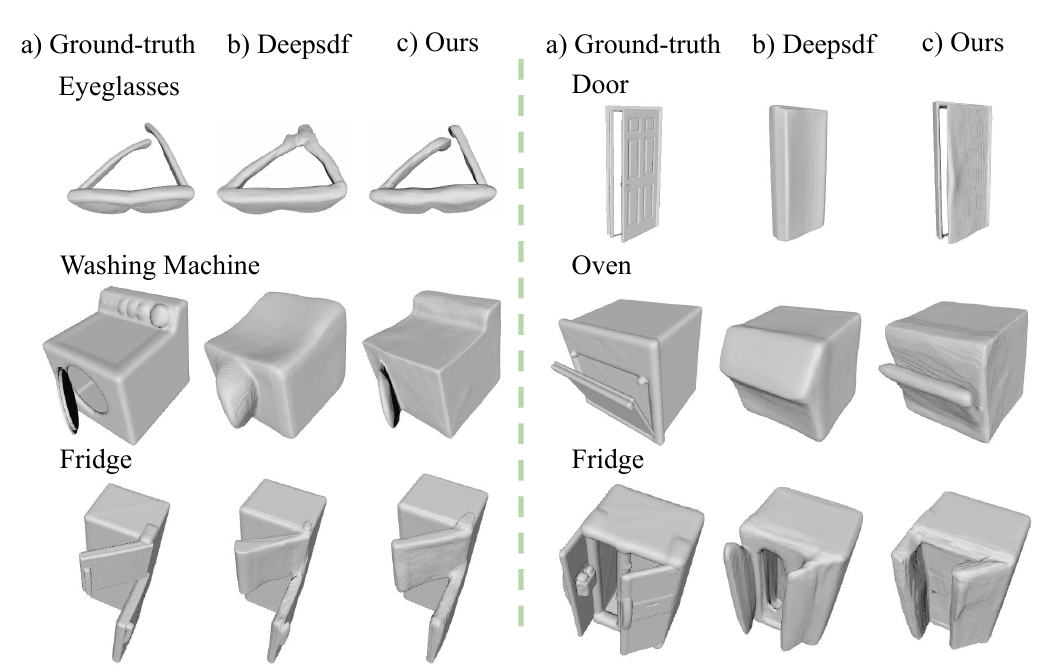}
    \caption{Reconstruction of test instances. The proposed method encodes better details whereas DeepSDF produces over-smoothed surfaces. Note that we model fridges with two different kinds of door types using a single network.}
    \label{fig:reconstruction}
\end{figure}
\begin{table}[t]
\tablestyle{2pt}{1.1}
\centering
\begin{tabular}{l|ccccccc} 
\multirow{3}{*}{} & Laptop & Stapler & Wash & Door & Oven & Eyeglasses & Fridge\\
\shline
DeepSDF~\cite{DBLP:conf/cvpr/ParkFSNL19}  & 0.35  & 3.73 & 4.29 & 0.61 & 5.33 & 1.63 & 0.80\\
Ours {\scriptsize(w/o TTA)} & 0.15 & 3.39 & 2.60 & 0.21 & 3.72 & 1.25 & 0.90 \\
Ours & \bf 0.13 & \bf 2.55 & \bf 2.13 & \bf 0.17 & \bf 1.83 & \bf 1.16 & \bf 0.69 \\
\end{tabular}
    \vspace{-0.10in}
\caption{Chamfer-L1 distance comparison for reconstruction. The proposed method yields smaller Chamfer-L1 distance.}
    \vspace{-0.10in}
\label{table:Reconstruction}
\end{table}

    \begin{table}[t]
    \tablestyle{2pt}{1.1}
    \centering
    \begin{tabular}{l|ccccccc} 
    \multirow{3}{*}{} & Laptop & Stapler & Wash & Door & Oven & Eyeglasses & Fridge\\
    \shline
    DeepSDF~\cite{DBLP:conf/cvpr/ParkFSNL19}  & 2.77 & 8.69 & 8.04 & 7.79 & 11.13 & 3.33 & 1.74\\
    Ours {\scriptsize(w/o TTA)}  & \bf \bf 0.17 & \bf 3.66  & \bf 3.81 & \bf 0.51 & \bf 6.83 &\bf 1.87  & \bf 0.69\\
    \end{tabular}
        \vspace{-0.10in}
    \caption{Chamfer-L1 distance comparison for interpolation.}
    \label{table:Interpolation}
        \vspace{-0.20in}
    \end{table}

\subsection{Experiment Setup}\label{exp:setup}
\textbf{Dataset.} For all experiments, the mesh models used are from the Shape2Motion dataset~\cite{DBLP:conf/cvpr/WangZSCZ019}. Shape2Motion is a large scale 3D articulated object dataset containing 2,440 instances. We select seven categories with sufficient number of instances per category, which are laptop, stapler, washing machine, door, oven, eyeglasses, and fridges. Each instance is articulated to $M$ poses. For each articulated shape, we follow \cite{DBLP:conf/cvpr/ParkFSNL19} to generate SDF samples. In addition to SDF values, we simultaneously generate part labels. Articulations for training are sampled in a systematic manner as detailed in the supplementary materials.

\textbf{Evaluation Metrics.} To evaluate the quality of generated shapes, Chamfer-L1 distance is used as the main evaluation metric, which is defined as the mean of an accuracy and a completeness metric. Following DeepSDF~\cite{DBLP:conf/cvpr/ParkFSNL19}, 30,000 points are sampled for each shape for evaluation and Chamfer-L1 distances shown in the paper are multiplied by 1,000. For joint angle estimation, we follow~\cite{DBLP:conf/cvpr/Li0YGAS20} to evaluate the average joint angle error in degrees for each joint. 


\textbf{Training and Inference.} For all experiments, shape codes are randomly initialized with $\mathcal{N}(0,0.01^2)$. $\lambda_{p}$ is set to $0.001$ and $\lambda_{\boldsymbol{\phi}}$ is set to $0.0001$. The predicted and ground-truth SDF values are clamped with 0.1 as discussed in ~\cite{DBLP:conf/cvpr/ParkFSNL19}. Network $f_{\theta}$ and shape codes are trained for 1,000 epochs. During inference, articulation codes are initialized to be the middle of the observed angle range. We optimize, the shape code, the articulation code, and the shape encoder each for 800 iterations with Adam Optimizer\cite{DBLP:journals/corr/KingmaB14}. The Marching Cubes algorithm~\cite{DBLP:conf/siggraph/LorensenC87} is applied to extract an approximate iso-surface given the predicted SDF values. More details about network designs are discussed in the supplementary materials.

\subsection{Reconstruction}\label{exp:recon}

We task our model with 3D reconstruction on the held-out data. For all methods, latent codes are first inferred using back-propagation given the observed samples, and then SDF values of sampled points are predicted by a forward pass.

\begin{figure}[t]
    \centering
    \includegraphics[width=0.9\linewidth]{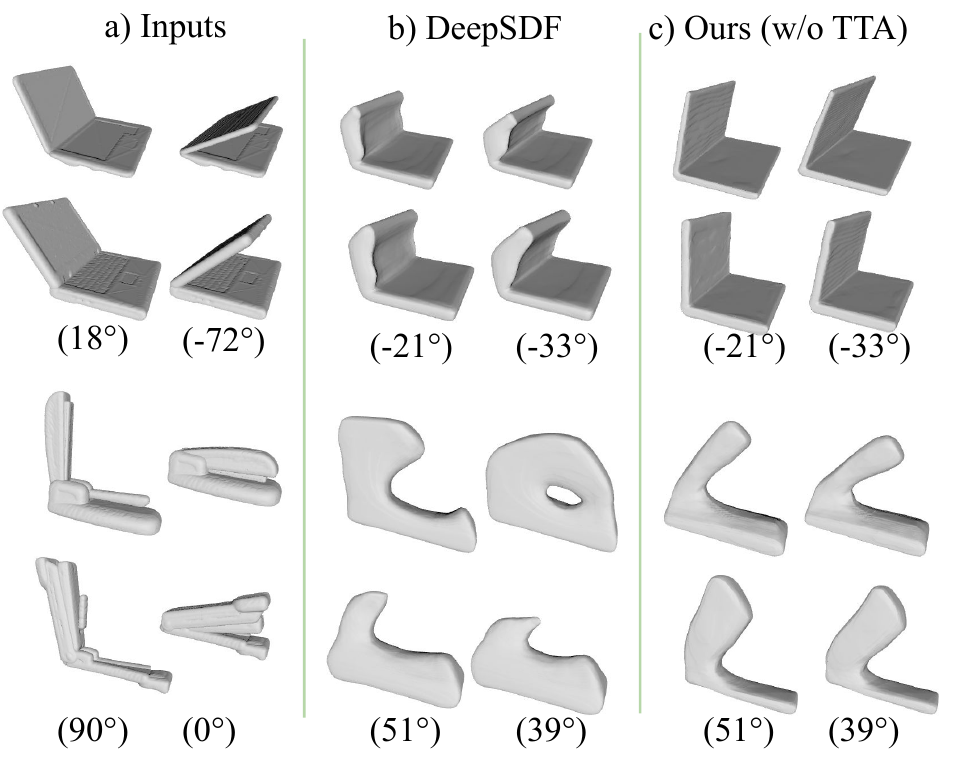}
    \vspace{-0.10in}
    \caption{Comparison for interpolation. Each row shows a different object. \textit{Inputs}, \textit{DeepSDF} , and \textit{Ours {\scriptsize(w/o TTA)}} results are shown from left to right respectively.}
    \label{fig:inter-hard}
\end{figure}

\begin{figure}[t]
    \centering
    \includegraphics[width=0.9\linewidth]{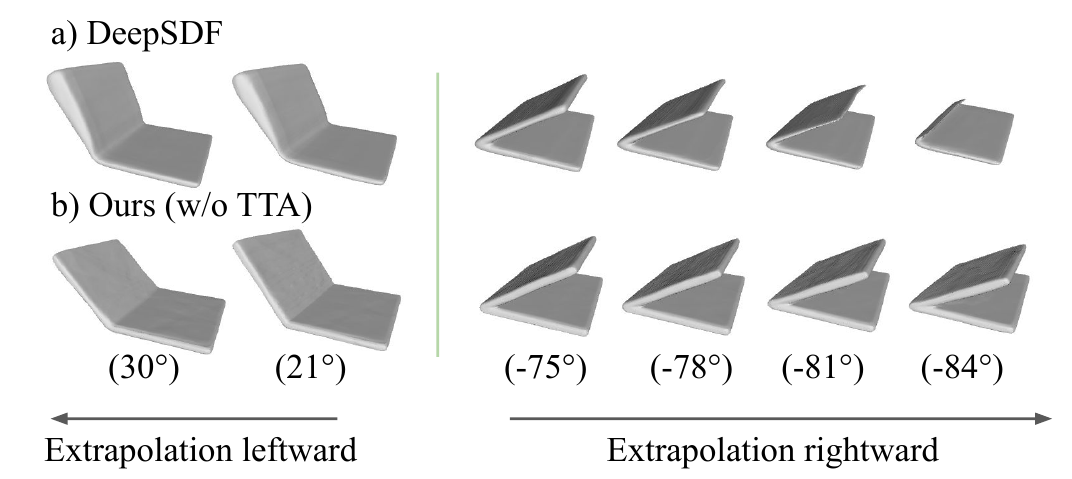}
    \vspace{-0.10in}
    \caption{Comparison for extrapolation. Each row is corresponding to the same instance extrapolated in two directions. The proposed method successfully generates shapes with joint angles beyond the range seen during training.}
    \label{fig:extrapolation}
    \vspace{-0.20in}
\end{figure}

    \begin{table*}[t]
    \tablestyle{8pt}{1.1}
    \centering
    \begin{tabular}{l|ccccccc} 
     & Laptop & Stapler & Washing  & Door  & Oven  & Eyeglasses & Fridge \\
    \shline
    \hspace{0.3cm}DeepSDF~\cite{DBLP:conf/cvpr/ParkFSNL19} \textit{(Interpolation)} & 2.77 & 8.69 & 8.04 & 7.79 & 11.13 & 3.33 & 1.74\\
    \hspace{0.3cm}Ours {\scriptsize(w/o TTA)} & 0.39~(1.39) &  3.77~(3.30) &  \bf 2.86~(7.10) &  0.73~(1.09) &  3.77~(7.08) &  2.48~(2.58) & 0.97~(3.47)\\ 
    \hspace{0.3cm}Ours & \bf 0.32~(1.59) & \bf 3.25~(3.53) &  3.01~(8.44) & \bf 0.53~(0.95) & \bf 2.58~(6.79) &  \bf 2.42~(2.84) & \bf 0.86~(4.19)\\ 
    \hline
    \hspace{0.3cm}Ours {\scriptsize(w/o TTA)} + part label & 0.32~(1.45) & 3.08~(3.66) & 2.16~(2.66) &  0.38~(1.04) &  5.19~(3.20) & \bf 2.03~(2.12) & 0.85~(3.69)\\ 
    \hspace{0.3cm}Ours + part label & \bf 0.29~(1.48) & \bf 2.48~(3.34) & \bf 1.96~(2.03) &  \bf 0.33~(1.67) & \bf 3.10~(2.98) &  2.16~(2.18) & \bf 0.64~(2.98)\\

    \end{tabular}
        \vspace{-0.10in}
    \caption{Chamfer-L1 distance comparison for shape synthesis. Joint angle estimation errors of the proposed method in brackets ($\cdot$). }
    \label{table:Generation}
        \vspace{-0.10in}
    \end{table*}

Quantitative evaluation shows the proposed method yields better results compared to DeepSDF for all classes, as shown in Table~\ref{table:Reconstruction}. Even though the proposed shape representation is more compact and represented with less parameters, quantitative results suggest that it reconstructs shapes with sharper boundaries. Our results show rich details compared to the over-smoothed results produced DeepSDF as visualized in Figure~\ref{fig:reconstruction}. We conjecture that, by introducing the articulation code and the shape code sharing, the proposed model can take advantage of multiple shapes articulated from the same instance to learn better shape priors. In addition, applying Test-Time Adaption helps the model fit the observation better. In Figure~\ref{fig:reconstruction}, we show that one single network is capable of learning priors for fridges with different in-class articulation patterns (door types).

\subsection{Interpolation and Extrapolation}\label{exp:inter}
The ability to interpolate and extrapolate between shapes is important for a good 3D articulated object representation. First, as the articulation space is continuous, training models to densely cover all angles is infeasible. Second, training models with densely sampled angles takes huge amount of training time, which makes it hard for real-world applications. In this section, given two articulated shapes of the same instance, we ask the models to output shapes with articulations in between or beyond.

As shown in Table~\ref{table:Interpolation}, we quantitatively demonstrate that the proposed model can reliably interpolate shapes. In both cases, the proposed method outperforms DeepSDF by a large margin. As visualized in Figure~\ref{fig:inter-hard}, the baseline generates unrealistic deformed shapes whereas the proposed method produces accurate shape estimations.

In the extrapolation setting on laptops, the proposed method also significantly outperforms DeepSDF. We compute the Chamfer-L1 distance for both methods. The proposed method achieves 0.280 whereas DeepSDF only gets 3.396. As illustrated in Figure~\ref{fig:extrapolation}, it is observed that the proposed method is capable of extrapolating beyond the angle range seen during training while DeepSDF fails to produce valid shapes.

The insight is that, even in cases where the two given shapes (the same instance) are with large articulation difference, the inferred shape codes of the proposed method are still close to each other in the high dimensional space. The difference between these two shapes is mainly captured by the  articulation codes. Therefore, the linear interpolation in both shape and articulation spaces still produces valid shapes. However, the baseline method learns a non-structured shape space and the interpolation result could be a random point in a high dimensional space without corresponding to any meaningful shape.

\begin{figure}[t]
    \centering
    \includegraphics[width=0.95\linewidth]{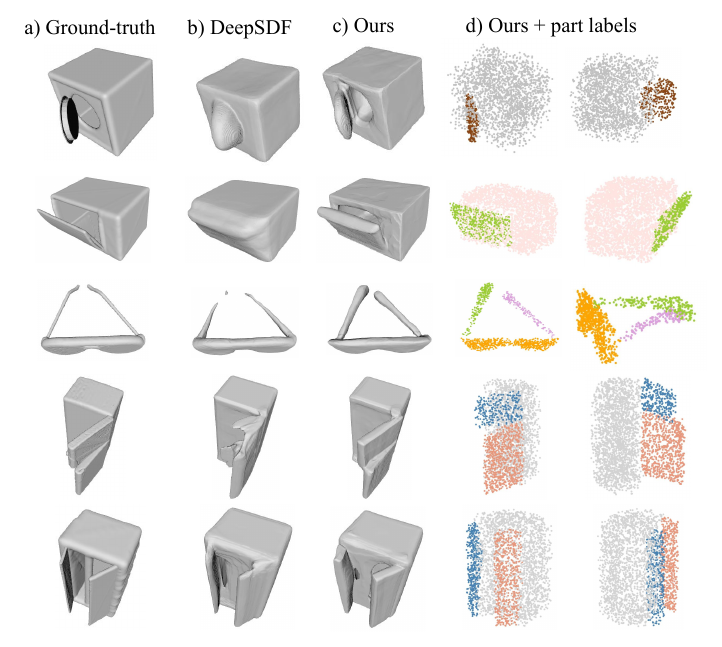}
    \caption{Shape synthesis and part prediction. From left to right: the ground-truth, \textit{DeepSDF interpolation}, \textit{Ours generation}, and the \textit{Ours + part labels} part prediction results.}
    \label{fig:part}
    \vspace{-0.20in}
\end{figure}

The detailed procedure is described as follows. For the proposed method, given two shapes, two set of codes are first inferred using the procedure shown in Section~\ref{method:Inference}. Then the inferred shape codes and articulation codes are separately interpolated. For the baseline, the target shape code is simply computed as a linear combination of the two obtained codes.

\begin{figure*}[t]
    \centering
    \includegraphics[width=0.95\linewidth]{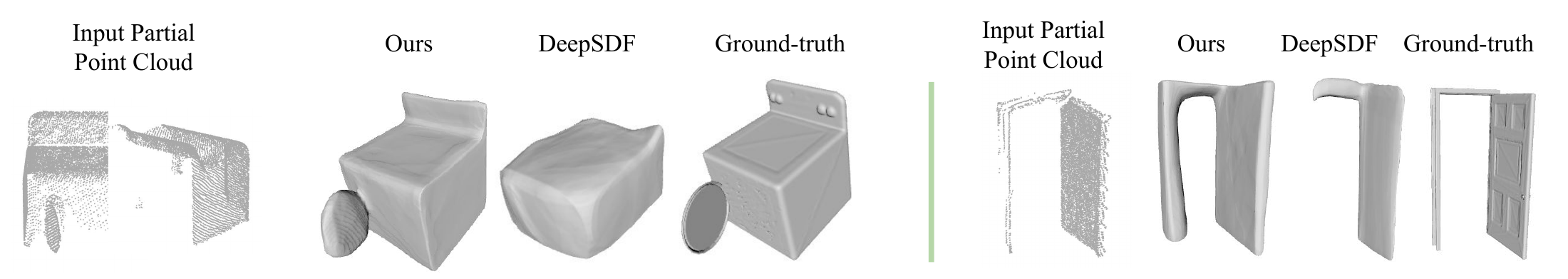}
    \vspace{-0.10in}
    \caption{Partial point clouds completion. For a given depth image visualized as point clouds, we show a comparison of shape completion from our method against DeepSDF and Ground-truth.}
    \label{fig:syn_pointcloud}
\end{figure*}

\begin{figure*}[t]
    \centering
    \includegraphics[width=\linewidth]{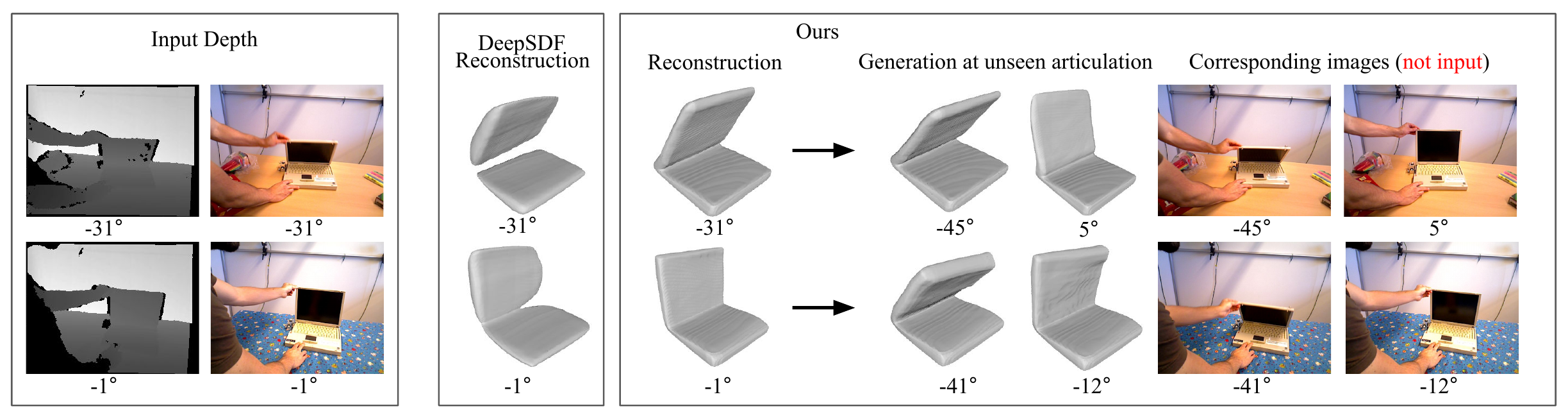}
    \caption{Test on real-world depth images. From left to right: \textit{Input depth}, \textit{DeepSDF reconstruction}, \textit{Ours reconstruction and generation}. Note that the \textit{Ours generation} are generated by only changing the articulation code. The shape code is inferred from the input depth of a laptop at a different articulation. RGB images and joint angles shown are only for visualization purposes and are not input to the model.}
    \label{fig:real_pointcloud}
    \vspace{-0.10in}
\end{figure*}

\subsection{Shape Synthesis and Part Prediction}\label{exp:gen}
One main advantage of our learned disentangled representation is its generation ability. We can easily control the articulation input to generate corresponding shapes of unseen instances with unseen joint angles. In this section, we study the quality of generated shapes using the proposed generation method in Section~\ref{method:Generation}.

To provide comparisons, we employ the DeepSDF Interpolation results described in Section~\ref{exp:inter} as baseline. Note that this is not a fair comparison as our method requires only one shape instead of two as for the baseline. Though relying on less information, the proposed method still yields much better results as shown in Table~\ref{table:Generation}. We demonstrate that applying Test-Time Adaption reduces the error further, indicating that Test-Time Adaption helps with inferring better shape while maintaining a disentangled representation. As visualized in Figure~\ref{fig:part}, note that one single network is capable of synthesize fridges with different articulation patterns.

One additional advantage of the proposed method is that joint angles can be estimated simultaneously. We quantitatively evaluate joint angle prediction errors in degrees, as shown in brackets in Table~\ref{table:Generation}. Results suggest that the proposed model can predicts joint angles accurately during the inference stage.

We also demonstrate that, if provided, part labels can further boost the performance. Models trained with part labels are denoted as \textit{Ours + part labels}. Part label prediction results are visualized in Figure~\ref{fig:part}.

\begin{table}[t]
    \tablestyle{3pt}{1.1}
    \centering
    \begin{tabular}{cl|cc|cc|cc} 
    \multirow{2}{*}{} & &\multicolumn{2}{c|}{Laptop} & \multicolumn{2}{c|}{Door} & \multicolumn{2}{c}{Washing}  \\
      & & Recon & Gen & Recon & Gen & Recon & Gen \\ [0.5ex]
    \shline 
    \multirow{3}{*}{2-view}
    & DeepSDF~\cite{DBLP:conf/cvpr/ParkFSNL19}  & 2.40 & 4.00 & 3.34 & 16.63 & 14.14 & 11.38\\
    & Ours {\scriptsize(w/o TTA)} &  0.75 &  1.16 &  0.62 &  0.83 & 7.63 & 10.03\\
    & Ours & \bf 0.61 & \bf 1.07 & \bf 0.61 & \bf 0.78 & \bf 5.31 & \bf 9.39\\
    \hline
    \multirow{3}{*}{1-view} 
    & DeepSDF~\cite{DBLP:conf/cvpr/ParkFSNL19}  & 3.31 & 6.75 & 5.76 & 20.24 & 14.08 & 13.05\\
    & Ours {\scriptsize(w/o TTA)} & \bf 2.19 & \bf 1.25 & \bf 0.80 & \bf 0.89 &  9.58 & 11.31\\
    & Ours &  2.25 &  1.55 &  0.86 & 1.38 & \bf 6.29 & \bf 9.52\\
    \end{tabular}
        \vspace{-0.10in}
    \caption{Chamfer-L1 distance comparison on partial point clouds. 1/2-view distinguishes the setting whether the partial point clouds are generated from one or two depth images.}
    \label{table:Shape completion}
    \vspace{-0.20in}
    \end{table}

\subsection{Test on Partial Point Clouds}\label{exp:synthetic depth}
We task the models with shape completion and generation on the partial point clouds from depth observations, which aims to inferring the shape code that best matches the partial point clouds and then generating to unseen articulations. We quantitatively demonstrate the proposed method outperforms the baseline significantly in Table~\ref{table:Shape completion}. Note that models are not trained on partial point clouds.

As 3D meshes are not provided in this scenario, we sample two points for each depth observation following DeepSDF~\cite{DBLP:conf/cvpr/ParkFSNL19}. We approximate the SDF values to be $\eta$ and $-\eta$ by perturb each of them $\eta$ distance away from the observed depth point along the computed surface normal direction. $\eta$ is set to be 0.025 in our experiments. Differently, as we deal with articulated objects, we do not sample points along the freespace-direction. 

As discussed in Section~\ref{exp:gen}, we employ the interpolation results of DeepSDF as a baseline for shape generation. We test the model on the point clouds generated from different number of depth images. As the number of views increases, we observe that applying Test-Time Adaption yields larger performance gain. We also visualize the reconstructed shapes of the one-view setting in Fig~\ref{fig:syn_pointcloud}, we observe that the proposed method encodes stronger shape prior and recovers shapes reliably given the partial point clouds input.

\subsection{Test on Real-world Depth Images}\label{exp:real depth}

    
    \begin{table}[t]
    \tablestyle{6pt}{1.1}
    \centering
    \begin{tabular}{l|cc} 
    \multirow{3}{*}{} & Reconstruction & Generation\\
    \shline
    DeepSDF~\cite{DBLP:conf/cvpr/ParkFSNL19}  & 4.65 & - \\
    Ours {\scriptsize(w/o TTA)}  & 2.53 & 5.09  \\
    Ours & \bf 0.76 & \bf 3.22 \\
    \end{tabular}
        \vspace{-0.10in}
    \caption{Chamfer-L1 distance comparison on real-world depth images. The Chamfer-L1 distance here is from ground-truth depth to reconstructed shape. DeepSDF is not able to generate new shapes.}
    \label{table:real depth}
        \vspace{-0.20in}
    \end{table}

We quantitatively show the proposed method generalizes better on real-world depth images, as shown in Table~\ref{table:real depth}. The RBO dataset~\cite{DBLP:journals/ijrr/Martin-MartinEB19} is a collection of 358 RGB-D video sequences of humans manipulating articulated objects, with the ground-truth poses of the rigid parts annotated by a motion capture system. We take laptop depth images from different sequences in the dataset and crop laptops from depth images by applying Mask R-CNN~\cite{DBLP:conf/iccv/HeGDG17} on the corresponding rgb images. We generate corresponding point clouds from real depth images following Section~\ref{exp:synthetic depth}, and then exploit the ground-truth pose to align the point clouds to the canonical space defined by Shape2motion dataset~\cite{DBLP:conf/cvpr/WangZSCZ019}.

In Table~\ref{table:real depth}, we show both reconstruction and generation results. Note both models are not trained on real-world depth images. Given a real-world depth image, we obtain its corresponding point clouds, input it to the model trained on synthetic data to reconstruct its 3D shape, and evaluate the reconstruction performance as the one-way Chamfer-L1 distance from ground-truth depth to reconstructed shape. Next, we take the shape code from the previous reconstructed shape and change the articulation code to output shapes at multiple unseen articulation. We take the real depth images at these new articulation and use the generated corresponding point clouds as the ground-truth to evaluate the generation performance. As visualized in Fig~\ref{fig:real_pointcloud}, the proposed model reliably synthesize shapes at unseen articulation whereas DeepSDF does not have the ability to generate shapes. Table~\ref{table:real depth} results suggest that applying Test-Time Adaption reduces the error further on both reconstruction and generation.

\section{Conclusions}
We propose Articulated Signed Distance Functions (A-SDF) to model articulated objects with a structured latent space. An Test-Time Adaptation inference algorithm is introduced to infer shape and articulation simultaneously. We experiment on seven articulated object categories from the shape2motion dataset~\cite{DBLP:conf/cvpr/WangZSCZ019}, and demonstrated improved shape reconstruction, interpolation, and extrapolation performance. Moreover, the method allows for controlling the articulation code to generate shapes for unseen instances with unseen joint angles. We also go beyond synthetic data and demonstrate the proposed method can reliably generate 3D shapes from real-world depth images from the rbo dataset~\cite{DBLP:journals/ijrr/Martin-MartinEB19}.

\vspace{1em}
{\footnotesize \textbf{Acknowledgements.}~This work was supported, in part, by grants from DARPA LwLL, iARPA (DIVA) D17PC00342, NSF IIS-1924937, NSF 1730158 CI-New: Cognitive Hardware and Software Ecosystem Community Infrastructure (CHASE-CI), NSF ACI-1541349 CC*DNI Pacific Research Platform, and gifts from Qualcomm and TuSimple.}

{\small
\bibliographystyle{ieee_fullname}
\bibliography{egbib}
}

\end{document}